# Optimizing Spatio-Temporal Information Processing in Spiking Neural Networks via Unconstrained Leaky Integrate-and-Fire Neurons and Hybrid Coding

Yang Liu[1,2,a], Huaxu He[1,2], Wenqian Cao[1,2], Yinghao Lin[1,2,a]

*1.Henan Key Laboratory of Big Data Analysis and Processing, Henan University, Kaifeng 475004, PR China*
*2. School of Computer and Information Engineering, Henan University, Kaifeng 475004, PR China*

**Abstract**

Spiking Neural Networks (SNN) exhibit higher energy efficiency compared to Artificial Neural Networks (ANN) due to their unique spike-driven mechanism. Additionally, SNN possess a crucial characteristic, namely the ability to process spatio-temporal information. However, this ability is constrained by both internal and external factors in practical applications, thereby affecting the performance of SNN. Firstly, the internal issue of SNN lies in the inherent limitations of their network structure and neuronal model, which result in the network adopting a unified processing approach for information of different temporal dimensions when processing input data containing complex temporal information. Secondly, the external issue of SNN stems from the direct encoding method commonly adopted by directly trained SNN, which uses the same feature map for input at each time step, failing to fully exploit the spatio-temporal characteristics of SNN. To address these issues, this paper proposes an Unconstrained Leaky Integrate-and-Fire (ULIF) neuronal model that allows for learning different membrane potential parameters at different time steps, thereby enhancing SNN' ability to process information of different temporal dimensions. Additionally, this paper presents a hybrid encoding scheme aimed at solving the problem of direct encoding lacking temporal dimension information. Experimental results demonstrate that the proposed methods effectively improve the overall performance of SNN in object detection and object recognition tasks. related code is available at https://github.com/hhx0320/ASNN.

*Keywords:* Spiking neural network, Direct encoding, Neuronal model, Spatio-temporal information;

## 1. Introduction

Spiking Neural Network (SNN), as the third-generation neural network models [1], exhibit remarkable characteristics including low power consumption, event-driven nature, and high similarity to biological neuromorphic systems. These features endow them with significant energy efficiency advantages on neuromorphic computing hardware platforms, and they have been successfully applied in fields such as image

[a]*Corresponding author.E-mail: linyh@henu.edu.cn(Yinghao Lin), ly.sci.art@gmail.com(Yang Liu)*

classification and robotic decision-making control [2]. The unique advantages demonstrated by SNN they possess broad application prospects across multiple domains [3-5]. In recent years, although the performance of some deep SNN has approached that of Artificial Neural Networks (ANN) [6], their performance still lags slightly behind when dealing with large datasets [7]. Additionally, SNN trained directly often struggle to match the performance of artificial neural networks with the same architecture [8], further highlighting the challenges faced in developing high-performance, low-latency SNN.

The neuron model serves as the core for processing temporal information in spiking neural networks, playing crucial roles in the network model, including but not limited to communicating temporal dimension information, implementing activation functions, and converting floating-point information into binary spikes. Currently, commonly used neuron models include LIF (Leaky Integrate-and-Fire) and IF (Integrate-and-Fire). Although these simple neuron models can meet the basic requirements for model operation, they adopt the same membrane potential parameters for input information at different time steps, which limits the full utilization of temporal information by the model. Previous improvements to neuron models have tended to complicate the models, increase learnable parameters, or increase computational load to facilitate gradient propagation, but these methods increase the cost of network training and the number of network parameters. Therefore, this paper proposes a method to enhance the network's ability to process temporal information using Time-wise and ULIF (Unconstrained Leaky Integrate-and-Fire) neurons without increasing the network's computational load.

In SNN, encoding input floating-point information into spike sequences is a crucial step that significantly influences network performance. Directly trainable SNN commonly adopt direct encoding methods [9-11]. The primary advantage of direct encoding lies in its ability to achieve excellent classification performance within a short period of time. However, its limitation is that, at each time step, the network receives the same feature map as input, which fails to fully exploit the unique spatio-temporal characteristics of SNN. In contrast, temporal encoding methods (with time to first spike encoding being used in this paper) convert input images into spike data with spatio-temporal dynamics. However, converting input images into spike sequences within short time steps often results in significant loss of useful information. Therefore, this paper proposes an innovative hybrid encoding method that combines the advantages of temporal encoding and direct encoding, aiming to significantly enhance the spatio-temporal dynamics of the direct encoding approach.

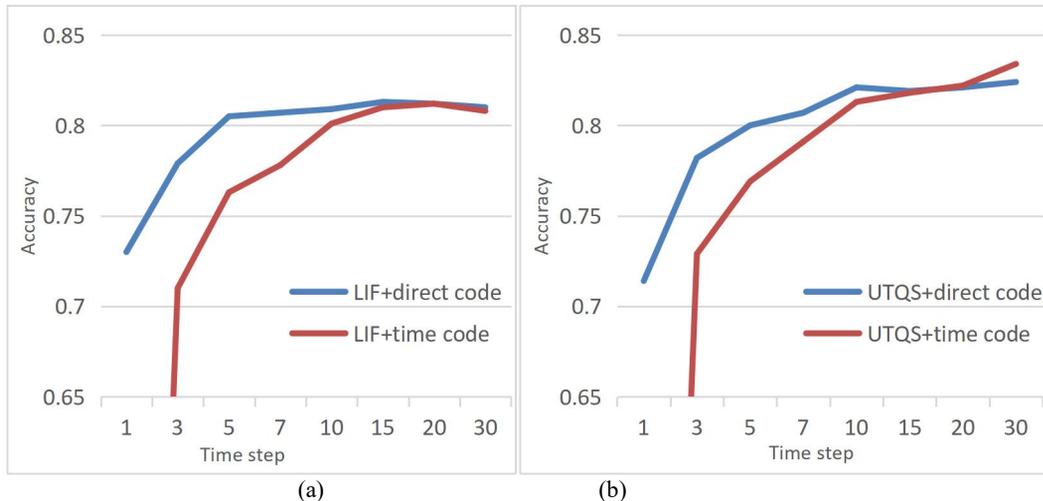

Figure 1, Comparison of direct encoding and time encoding. using the network architecture of LeNet, on the cifar10 dataset, the training results are obtained by using time steps 1, 3, 5, 7, 10, 15, 20 and 30 respectively. Figure (a) shows the comparison between direct encoding and time encoding of LIF neurons, and figure (b) shows the comparison between direct encoding and time encoding when using UTQS.

In summary, SNN inherently possess the advantage of processing temporal information. However, current network model structures within SNN are unable to fully exploit this advantage, leading to difficulties in effectively mining temporal information from data during actual training. As shown in Figure 1(a), when using the LIF neuron model, the performance of temporal encoding consistently falls below that of direct encoding. This deficiency in temporal information processing capabilities compels most researchers to prefer direct encoding, which converts input images into representations devoid of temporal information, when utilizing directly trained SNN. Therefore, enhancing the network model's ability to apply temporal information and adopting encoding methods with more pronounced spatio-temporal dynamics are crucial factors in improving SNN performance.

The main contributions of this paper are summarized as follows:

- Proposing the ULIF Neuron Model: This model extends the learnable membrane potential constant range from (0, 1) to the real number domain, and combines it with the Time-wise method and output quantization scheme to significantly enhance the network's ability to process and apply temporal information. For ease of description, when these three schemes are used together, we refer to them collectively as UTQS (Unconstrained Time-wise Quantify Spiking). The experimental results show that when the time step reaches 30, the performance of UTQS using time encoding is significantly better than that using direct encoding, as shown in Figure 1 (b).
- Proposing the Hybrid Encoding Strategy: This strategy combines direct encoding with temporal encoding, and provides this hybrid encoding as input to the network. Experimental results indicate that the performance of this strategy is superior to both direct encoding and temporal encoding, regardless of whether the time step is low or high.
- Experimental Verification and Performance Analysis: A series of experiments were conducted to verify the effectiveness of the proposed methods. In comparative experiments, the methods proposed in this paper exhibited significantly better performance than the baseline model. Additionally, ablation experiments were performed for object detection and object recognition tasks, further demonstrating the characteristics of the proposed methods. Under the same network computation, the proposed methods offer significant performance advantages.

## 2. Related Works

*2.1. Training of Spiking Neural Networks*

Currently, the mainstream approaches for training deep SNN are mainly divided into two strategies: ANN to SNN [12-17], and directly training SNN [18-23]. The former approach approximates the behavior of ANN through the average firing rate of spikes in SNN, with its performance relying on the original ANN. However, this method often suffers from performance loss and high latency during the conversion process. In contrast, direct training faced challenges early on due to the non-differentiability of the spiking function, making it difficult to apply traditional backpropagation algorithms. With the application of surrogate gradients and the proposal of the Spatio-Temporal Backpropagation (STBP) method, direct training has gradually become one of the mainstream strategies for training SNN. This paper adopts the direct training method to optimize and experimentally study SNN.

Deep residual structures are key factors in improving network performance. In directly trained SNN, there are two main implementations of residual design: SEW-ResNet accumulates the 0,1 spiking signals generated after neuronal processing, while MS-ResNet accumulates the floating-point input current data before entering the neurons. Both methods have successfully achieved effective training of neural networks exceeding 100 layers [11, 24], and the experimental section of this paper includes relevant studies on both residual connection methods.

*2.2. Neuron Model*

The neuron model is a crucial component of the SNN model. The primary behavioral processes of spiking neurons can be described as: receiving current input, firing spikes based on the current membrane potential, and resetting the membrane potential. Currently, commonly used neuron models in SNN include the LIF [25-28] and IF [29] models. Compared to IF neurons, LIF introduces a membrane potential decay constant, where this decay represents the leakage mechanism of the potential, aiding in the stable convergence of neuronal dynamics and stabilizing the potential towards the resting potential [30].

In recent years, optimizations and improvements to neuron models have primarily focused on two aspects: making various parameters within neurons learnable and increasing the complexity of neuron models. For instance, PLIF first utilized learnable membrane potential decay constants trained alongside the network model [31]. DIET_SNN enhanced the model by adding new learnable parameters, making the spike firing threshold also a learnable component of the model [32]. GLIF introduced multiple sets of hyperparameters to balance three different behavioral characteristics within neurons, thereby enhancing neuronal dynamics [30]. DTAM+DSGM employed adaptive input currents and made different adjustments based on the sign of the input current to improve network performance [33]. CLIF optimized the process of network gradient propagation by modifying neuronal behavior and adding additional complementary potentials [34].

Among the aforementioned neuron models, PLIF improved network performance without increasing the computational load of network inference by modifying fixed values in LIF neuron parameters to learnable ones. The remaining neuron models, however, enhanced network performance by increasing the complexity of the neuron model. Compared to these methods, the ULIF proposed in this paper makes simple modifications to PLIF and combines the use of the Time-wise method. ULIF achieves significant performance improvements with an equivalent computational load for network inference as LIF and PLIF.

*2.3. Differences in SNN Output*

The final layer of SNN models typically employs average membrane potential or firing rate for prediction. Zhao et al. point out that this approach does not consider the impact of SNN output distribution across different time steps on performance. For constructing a stable and high-performance SNN, reducing the variance in SNN output distribution across different time steps is crucial [35].

The TET method avoids falling into local minima by optimizing presynaptic inputs at each time instant, rather than directly optimizing the average of outputs across different time steps in the final layer of the network [36]. Both methods underscore the importance of SNN output optimization. This paper proposes a simple optimization scheme for network output, namely output quantization, focusing on addressing the core issue of different time steps' inputs having varying importance and properties.

*2.4. Spike Encoding*

Spike encoding aims to convert input floating-point data into 0/1 spike-form data with a temporal dimension. The direct encoding method has received widespread attention since its proposal and currently dominates directly trained spiking neural networks (SNN) [9, 37, 38]. This encoding scheme first processes the input image through a convolutional layer, then repeats the obtained feature map multiple times along the temporal dimension, and finally inputs it into neurons to generate the spikes required by the SNN. However, this method uses the same feature map at each time step, failing to fully exploit the spatio-temporal characteristics of SNN.

In contrast, time to first spike encoding inputs different spike data at different time steps, effectively enhancing the spatio-temporal dynamics of the encoded spike data. Since each pixel in the input image can only release a spike once during the entire time period, this encoding method significantly reduces energy consumption during network operation [39]. However, time to first spike encoding requires a longer number of time steps to represent accurate information, as shown in Figure 2. Although time to first spike encoding has been widely used in the field of ANN to SNN [40, 41], it is rarely applied in directly trained SNN. This paper proposes a hybrid encoding scheme that combines direct encoding with temporal encoding and feeds it into the network, effectively improving network performance by enhancing the spatio-temporal dynamics of direct encoding.

Additionally, to address the issue of large networks being unable to use higher time steps for training, this paper adopts Weighted Phase Encoding to shorten the time steps required for encoding. Weighted Phase Encoding assigns different weights to spikes at different time steps, exhibiting overall performance similar to Rate Coding, where higher firing rates often represent larger values, while also retaining the characteristic of time to first spike encoding where earlier time steps represent larger values [42].

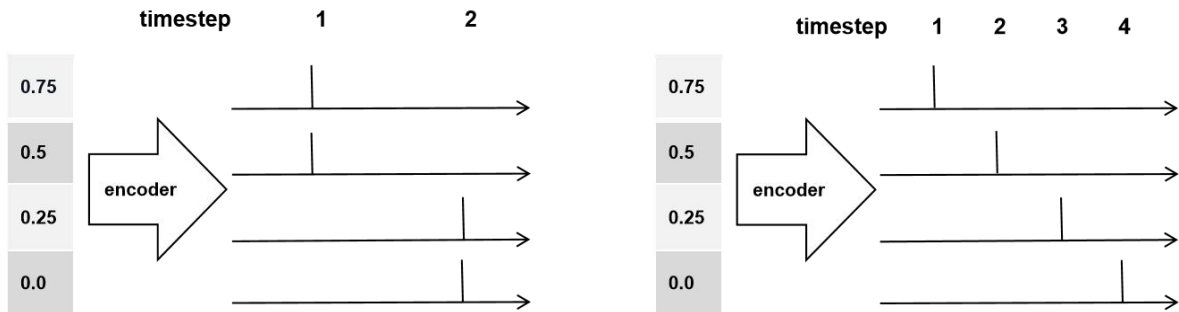

Figure 2, illustrates the legend for time to first spike encoding. The left figure shows two time steps, while the right figure shows four time steps. The accuracy of input information increases with the increase in the number of time steps.

## 3. Methods

### 3.1. Unconstrained Leaky Integrate-and-Fire

In SNN, the neuron functions analogously to the activation function in ANN, primarily responsible for converting input floating-point signals into discrete 0/1 spiking signals. Although neurons can vary in design complexity, overly complex models are often constrained by current computer hardware performance, especially when the network depth increases, leading to a sharp rise in computational overhead and complexity. Therefore, in practical applications, the use of complex neuron models is relatively infrequent. Currently, the formula for the commonly used neuron model in SNN is as follows:

$$H[t] = f(V[t-1], X[t]) \quad (1)$$
$$S[t] = \Theta(H[t] - V_{th}) \quad (2)$$
$$V[t] = H[t](1 - S[t]) + V_{reset}S[t] \quad (3)$$

Where $X[t]$ represents the input current at time step $t$. When the membrane potential $H[t]$ exceeds a threshold value $V_{th}$, the neuron fires a spike. $\Theta(V)$ is the Heaviside step function, which equals 1 for $V \geq 0$ and 0 otherwise. $V[t]$ denotes the membrane potential after a spiking event; if no spike is generated, it remains equal to $H[t]$, otherwise, it is set to the reset potential. The functional descriptions of the IF and LIF models can be formulated in Equations (4) and (5), respectively, as follows:

$$H[t] = V[t-1] + X[t] \quad (4)$$
$$H[t] = V[t-1] + \frac{1}{\tau}(X[t] - V[t-1] - V_{reset}) \quad (5)$$

Among them, τ represents the membrane time constant. Equations (2) and (3) describe the generation and resetting process of spikes, which are consistent across all types of spiking neuron models.

When utilizing encoding schemes with rich spatio-temporal dynamics, the data input at each time step carries different importance and unique information. However, traditional LIF and IF neurons often rely on fixed constants when processing membrane potential and input current, leading to identical or similar processing methods for input data from different time steps. This limits the network's ability to differentiate between input data from different time steps. Therefore, this paper uses Time-wise method, which allows each time step's input current and membrane potential decay constant to have independent learnable parameters. This enables the network to more autonomously choose whether to fire a spike at any moment, rather than following a fixed spiking pattern, as shown in Figure 3.

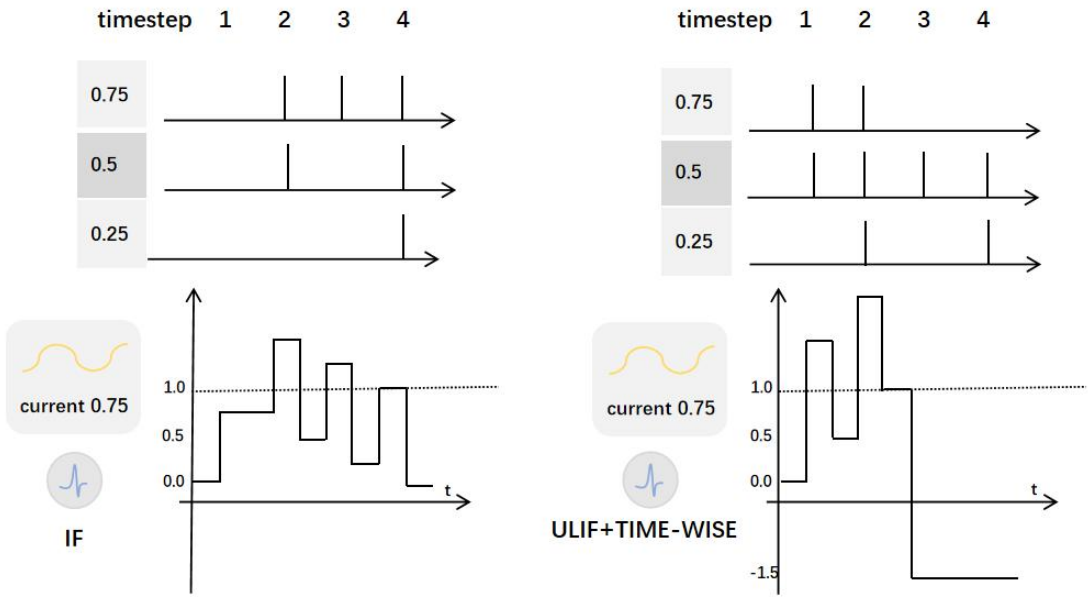

Figure 3. Firing Patterns of Spiking Neurons: The two diagrams on the left illustrate the firing patterns of IF neurons, while the two diagrams on the right depict the firing patterns of ULIF+TIME-WISE neurons. The upper diagrams show the neuronal firing patterns when the input current at each time step is fixed at 0.75, 0.5, and 0.25, respectively. The lower diagrams demonstrate the complete process of neuronal spiking with a constant input current of 0.75. During the reset phase, a soft reset mechanism is employed, where the membrane potential decreases by one unit after the neuron fires a spike. In the left diagrams, the membrane potential decay constants are set to (1, 1, 1, 1), and the input current constants are also set to (1, 1, 1, 1), constituting a fixed firing pattern. In contrast, in the right diagrams, the membrane potential decay constants are set to (1, 1, -3, 2), and the input current constants are set to (2, 2, 2, 2), representing just one possible firing pattern within the ULIF+Time-wise framework.

In existing neuronal models, the weight parameters of input currents are often tied to the membrane potential decay constant or fixed at a value of 1. This design lacks sufficient biological plausibility and limits the dynamic characteristics of neurons. To overcome this limitation, this paper separates the membrane potential decay constant and input currents, introducing two independent adjustable parameters so that their sum is no longer fixed at 1. Furthermore, in traditional implementations, the membrane potential decay constant is typically constrained to the range [0,1] using a Sigmoid function. In our experiments, we extend the range of the decay constant to the entire real number domain and demonstrate through experimental results that this approach offers performance superior to that of LIF neurons and equivalent to that of PLIF neurons. Combining these two methods, this paper proposes the ULIF neuronal model, which greatly expands the dynamic characteristics of neuronal models. This model is represented by Equation (6).

$$H[t] = l^t V[t-1] + i^t X[t] \qquad (6)$$

The incorporation of Time-wise in Equation (8) allows for learnable parameters that differ at each time step. Where $l^t$ represents the membrane potential decay constant at time step $t$, and $i^t$ is the input current constant at time step $t$. When both $l^t$ and $i^t$ are constantly equal to 1, Equation (8) degenerates into the IF neuron model. Specifically, when $l^t + i^t = 1$ and are all greater than 0, Equation (8) is consistent with the LIF neuron model.

*3.2. Output Quantization*

Traditional SNN typically add and average the output results directly to calculate the loss, as shown in equation (7), where $O(t)$ represents the output result of the last layer of the network, $y$ represents the target label, and the loss calculation uses cross entropy loss.

$$\mathcal{L}_{SNN} = \mathcal{L}_{CE}\left(\frac{1}{T}\sum_{t=1}^{T} O(t), y\right) \quad (7)$$

Addressing the issue of varying importance of input information at different time steps in processing temporal data with SNN, this paper proposes a simple and effective solution. Specifically, since input information at different time steps may vary in both its informational content and its significance, traditional methods that directly average the outputs of all time steps may lead to information loss or misinterpretation. To overcome this limitation, we introduce a set of learnable parameters that dynamically adjust the network's output at each time step. The specific operation involves multiplying the network's output at each time step by its corresponding learnable parameter and then summing these weighted outputs to obtain the final output. In this way, the network can quantify and evaluate the influence of input spiking data at each time step on the final output. For example, if early time steps contain less information or higher noise, the network can choose to multiply the outputs of the first few time steps by smaller weights (even close to 0), effectively ignoring their influence. When calculating the loss, we use Equation (8) to represent it, where $s^t$ denotes the network output weight at time step $t$.

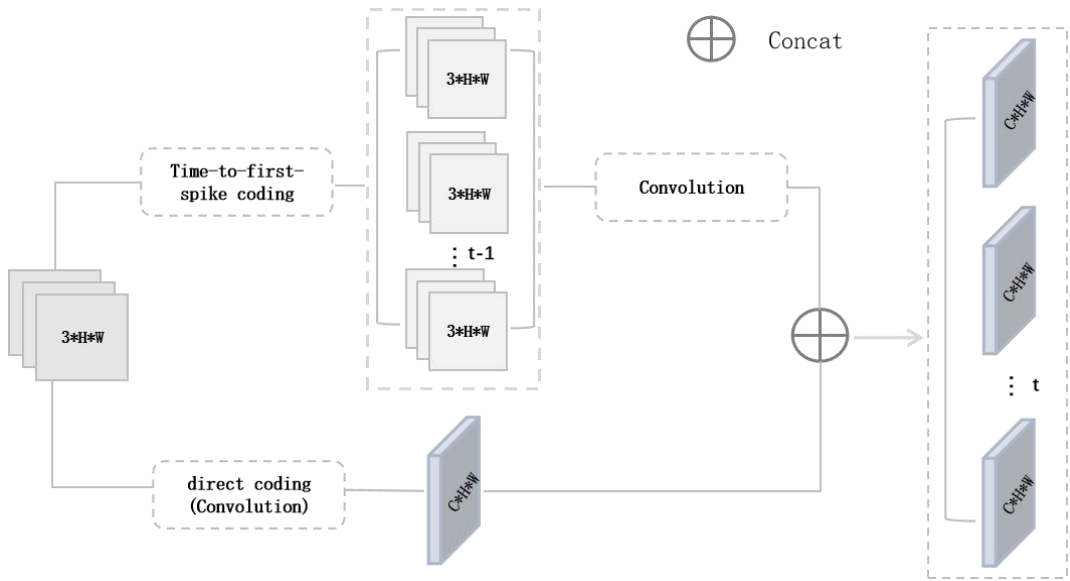

Figure 4, schematic diagram of Hybrid encoding. Input an image of 3 * H * W.

$$\mathcal{L}_{ASNN} = \mathcal{L}_{CE}\left(\frac{1}{T}\sum_{t=1}^{T} s^t O(t), y\right) \quad (8)$$

*3.3. Hybrid Encoding*

Hybrid encoding refers to an encoding scheme that combines direct encoding with other encoding methods. When hybridized with time to first spike encoding encoding, this scheme processes inputs at the first layer of the network. Specifically, the spiking sequences encoded by time to first spike encoding and the original images are processed through their respective convolutional layers. Subsequently, these processed feature maps are concatenated along the temporal dimension, as shown in Figure 4. In the experiments conducted in this paper, we attempted to concatenate the direct encoding either before or after the time to first spike encoding and found no significant difference in the final results. This hybrid encoding approach compensates for information potentially lost during time to first spike encoding through direct encoding, thereby enhancing the network's overall performance and demonstrating good classification performance.

## 4. Experiments

This section is mainly divided into four parts. In the first part, we combined the method proposed in this paper with deep networks such as Spikformer and MS-ResNet18 [43], and compared it with existing work. In the second part, using a completely consistent LeNet network structure, we conducted comparative experiments between our method and ANN as well as traditional SNN. The results further validated the effectiveness of our method. In the third part, we continued to use the LeNet network to conduct in-depth exploration of the characteristics of ULIF neurons and hybrid encoding. In the fourth part, to address the issue of experimental equipment being unable to meet the operational requirements of large-scale datasets, we added additional object detection tasks to increase the number of experimental samples. In this part, we used the YOLOv5n [44] network structure and modified its internal structure to a spiking manner, while adopting the same connection method as SEW-ResNet. Experiments with different model structures and tasks further validated the effectiveness of our method.

*4.1. Comparison with Existing Work*

In comparing with existing work, this paper adopts two widely used deep SNN architectures, MS-ResNet and Spikformer, as baseline models. To ensure fairness and accuracy of the experiments, the various parameter settings during network training are consistent with the source code of MS-ResNet and Spikformer. The experimental datasets used are CIFAR10 and CIFAR100, which are widely employed in classification tasks [45].

Table 1 provides detailed accuracy performances of different methods on the CIFAR10 and CIFAR100 datasets, revealing the effectiveness of the proposed method in this paper. Specifically, for the CIFAR10 dataset, when using 5 timesteps, the Spikformer model incorporating the proposed method achieves a significant performance improvement, with the accuracy increasing from 94.44% to 95.72% compared to the baseline model. Meanwhile, MS-ResNet combined with the proposed method achieves an accuracy of 96.84% at 6 timesteps, significantly outperforming its baseline model.

On the more challenging CIFAR100 dataset, when using 5 timesteps, the Spikformer model combined with the proposed method improves the accuracy from 75.74% to 77.35%. As the number of timesteps increases to 6, the accuracy further increases to 78.19%. Similarly, MS-ResNet combined with the proposed method achieves an accuracy of 78.33% at 6 timesteps, significantly better than the baseline model.

These results fully demonstrate the wide applicability of the proposed method across different network architectures and datasets, proving the significant effectiveness of the proposed hybrid weighted phase encoding technique in enhancing the performance of spiking neural networks.

Table 1 compares with existing work results, where * represents the results achieved on open source code.

| Dataset | Methods | Architecture | Coding | Time steps | Accuracy |
|---|---|---|---|---|---|
| CIFAR10 | Spikformer | Spikformer-4-256* [48] | direct | 5 | 94.44 |
| | | | | 6 | 95.19 |
| | STDP-tdBN | ResNet19[49] | direct | 4 | 92.92 |
| | SLT-TET | ResNet19[50] | direct | 6 | 95.26 |
| | MS-ResNet | ResNet18[51] | direct | 6 | 94.92 |
| | GLIF | ResNet19[30] | direct | 6 | 95.03 |
| | ANN | ResNet19 [48] | - | 1 | 94.97 |
| | Our methods | Spikformer-4-256 | hybrid weight phase | 5 | **95.72** |
| | | | | 6 | 94.91 |
| | | MS-ResNet18 | hybrid weight phase | 6 | **96.84** |
| CIFAR100 | Spikformer | Spikformer-4-256*[48] | direct | 5 | 75.74 |
| | | | | 6 | 76.69 |
| | STDP-tdBN | ResNet19[49] | direct | 4 | 70.86 |
| | SLT-TET | ResNet19[50] | direct | 6 | 74.87 |
| | MS-ResNet | ResNet18[51] | direct | 6 | 76.41 |
| | GLIF | ResNet19[30] | direct | 6 | 77.28 |
| | ANN | ResNet19 [48] | - | 1 | 75.35 |
| | Our methods | Spikformer-4-256 | hybrid weight phase | 5 | 77.35 |
| | | | | 6 | **78.19** |
| | | MS-ResNet18 | hybrid weight phase | 6 | **78.33** |

## 4.2. Comparison with ANN of the Same Structure

In this experimental section, we primarily adopted the LeNet network model as the benchmark model. The primary reason for selecting LeNet lies in its lightweight nature, which allows us to flexibly adopt longer time steps when exploring the impact of different time-step encoding schemes on network performance, thereby enabling a more comprehensive evaluation of the effectiveness of the encoding schemes.

The experimental results indicate that, under the same network structure, the method combining hybrid time code with UTQS exhibits higher classification performance compared to the method using direct code with LIF neurons. Specifically, on the CIFAR-10 and CIFAR-100 datasets, when the time steps are set to 10 and 15, respectively, the classification accuracy of the proposed method in this paper reaches a level comparable to that of ANN. In the case of short time steps, the performance of UTQS+hybrid time code is similar to that of LIF+direct code. However, as the time steps increase, the performance gap between UTQS+hybrid and LIF+direct code gradually widens, ultimately resulting in the classification performance of SNN surpassing that of ANN and significantly outperforming LIF+direct code, as illustrated in Figure 5.

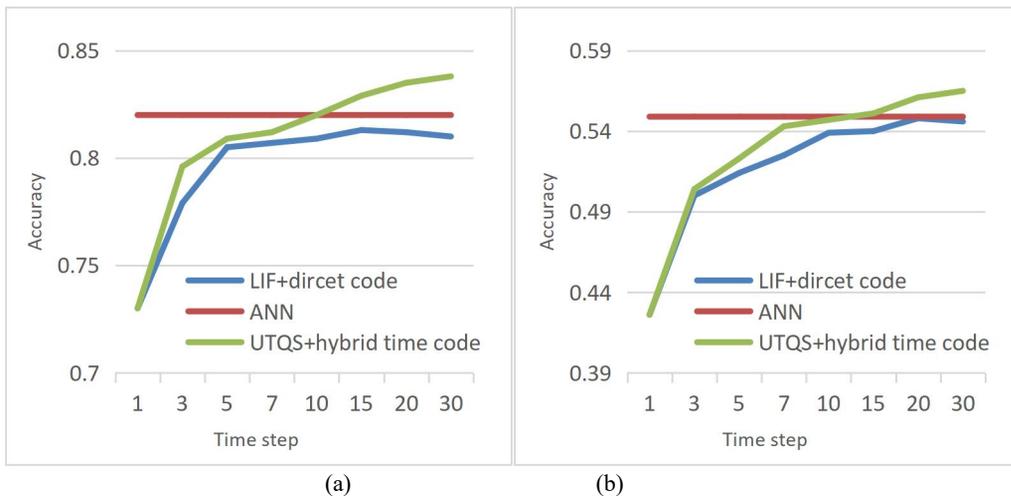

(a)      (b)

Figure 5. Comparison Diagram with ANN of the Same Structure. This figure presents the comparison of classification performance when using the LeNet network structure, compared with an Artificial Neural Network (ANN) of the same structure. Figure (a) uses the CIFAR-10 dataset, while Figure (b) uses the CIFAR-100 dataset. The three curves in the figures represent the trends of classification accuracy over time steps for ANN, LIF+direct code, and UTQS+hybrid time code, respectively. It should be noted that when the time step is 1, hybrid time code is equivalent to direct code, i.e., they have the same encoding method at this point.

*4.3. Verification Experiments on Method Characteristics*

The primary objective of the verification experiments presented in this section is to delve into the unique properties of the neuron model and encoding method proposed in this paper. The experiments continue to employ the LeNet network architecture and utilize the CIFAR-10 dataset for testing. The experiments focus on the following core aspects: 1)Exploring the Effectiveness of UTQS in Enhancing the Accuracy of Time Encoding and Direct Encoding. 2)Analyzing the Correlation Between Hybrid Time Encoding and Time Encoding, As shown in Figure 6.

The results of the method characteristic experiments revealed several key findings:

- **UTQS Effectively Enhances the Accuracy of Time Encoding and Direct Encoding:** Specifically, under the condition of 30 time steps, the direct encoding method employing UTQS exhibited a 0.4% accuracy improvement compared to an ANN of the same architecture. Meanwhile, time encoding showed a 1.4% accuracy improvement compared to the same-architecture ANN. Notably, when using traditional LIF neurons, the performance of both encoding methods was lower than that of the ANN of the same structure.
- **The Impact of UTQS on Direct Encoding is Less Than That on Time Encoding:** This significant difference results in time encoding surpassing direct encoding in accuracy after a sufficient number of time steps. This finding suggests that the application of UTQS in time encoding may hold greater advantages.
- The Accuracy of Hybrid Time Encoding Gradually Approaches That of Time Encoding as Time Steps Increase:

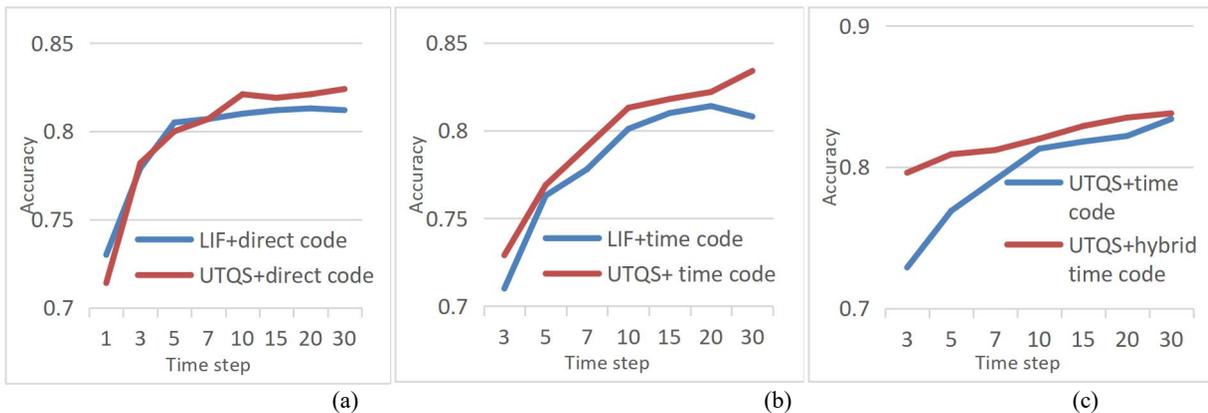

Figure 6 shows a comparison of method characteristics. Figure (a) shows a comparison between UTQS and LIF neurons for direct encoding, Figure (b) shows a comparison between UTQS and LIF neurons for direct encoding, and Figure (c) shows a comparison between time encoding and hybrid time encoding.

*4.4. Object detection related*

In this subsection, we conducted independent ablation experiments on the ULIF, Time-wise, and output quantization method within the UTQS framework, using YOLOv5n as the base model. The object detection dataset employed was the widely used PASCAL VOC. To accommodate the need for running high timesteps

YOLO networks on devices, we resized the input images to 224×224. The network performance evaluation indicators are mAP@0.5. In the neuron comparison experiments, we only compared ULIF with LIF neurons and PLIF neurons of equivalent computational complexity. Although adopting more complex neurons may bring performance improvements, this would also significantly increase training overhead and computational load of the network model.

Regarding the experiments on neuron and Time-wise quantization method combinations, the results indicated that the ULIF+Time-wise combination exhibited superior performance compared to other neuron combinations. The performance of other neuron models varied across different timesteps, as shown in Table 2. It is worth noting that all experimental results in Table 2 employed output quantization and direct encoding methods.

Regarding the experiments on neuron and Time-wise method combinations, the results indicated that the ULIF+Time-wise combination exhibited superior performance compared to other neuron combinations. The performance of other neuron models varied across different timesteps, as shown in Table 2. It is worth noting that all experimental results in Table 2 employed output quantization and direct encoding methods.

Table 2, Neuron Model Ablation Experiment,

| Timesteps | 2 | 4 | 6 | 8 | 10 | 15 |
|---|---|---|---|---|---|---|
| LIF | 0.341 | 0.418 | 0.475 | 0.507 | 0.533 | 0.528 |
| PLIF | 0.341 | 0.428 | 0.473 | 0.514 | 0.525 | 0.545 |
| ULIF | 0.352 | 0.422 | 0.485 | 0.507 | 0.512 | 0.55 |
| PLIF+Time-wise | 0.340 | 0.425 | 0.488 | 0.510 | 0.530 | 0.53 |
| ULIF+Time-wise | **0.357** | **0.455** | **0.497** | **0.520** | **0.542** | **0.570** |

Next, we replaced the output quantization method in the UTQS network structure with the traditional method of averaging the outputs, and recorded the experimental results. The specific experimental results are shown in Table 3. By comparing these results, we can gain a more intuitive understanding of the impact of the output quantization method and the averaging method on network performance.

Table 3 Output quantification method ablation experiment

| Timesteps | 2 | 4 | 6 | 8 | 10 | 15 |
|---|---|---|---|---|---|---|
| | 0.357 | 0.455 | 0.497 | 0.520 | 0.542 | 0.570 |
| output quantization->mean | 0.345 | 0.44 | 0.48 | 0.519 | 0.536 | 0.562 |

To further explore the impact of encoding methods on performance, we adopted 15 timesteps in the UTQS and replaced direct encoding with hybrid time encoding. The experimental results are presented in Table 4, which shows the impact of this change on model performance.

Table 4: Experimental ablation of encoding methods for object detection

| Method | Coding | Time steps | mAP@0.5 |
|---|---|---|---|
| ANN | - | - | 0.609 |
| SNN(UTQS) | Direct | 15 | 0.570 |
| SNN(UTQS) | Hybrid | 15 | **0.577** |

## 5. Analysis and Discussion

### 5.1. Analysis of UTQS

For neurons, compared to the LIF and IF neuron models that use fixed constants, the Time-wise method makes the dynamic characteristics of neurons more complex. Under the condition of constant current input, when the membrane potential decay constant at a certain time point is greater than that at the previous time point, the speed at which the membrane potential returns to the resting potential slows down. This results in the average membrane potential at the subsequent time step being more likely to remain at a higher level compared to the previous time step in adjacent time steps. This higher membrane potential state increases the likelihood of triggering a spike at the current time point, leading to a reduction in the firing threshold at the current time point compared to the previous time point, as shown in Equation (8). When $l^t > l^{t-1}$, it means that the firing threshold at time t is lower than that at time t-1. Conversely, if the membrane potential decay constant decreases compared to the previous time point, it may lead to an increase in the firing threshold.Another reason why our method directly affects the firing threshold is that when attempting to introduce the learnable firing threshold from the DIET-SNN into the UTQS, it is difficult to observe significant changes in network performance, and it may even result in some performance loss. In addition, the dynamic variation of the firing threshold is an inherent attribute of biological spiking neurons.      Unbinding the current input constant from the membrane potential decay also has a significant impact on the dynamic characteristics of neurons. Although the current input constant seemingly exhibits similar characteristics to the membrane potential decay constant in controlling the firing threshold—i.e., a rising current input constant over time implies a decreasing firing threshold, and a falling current input constant over time implies an increasing firing threshold — in experimental observations, the authors found that these two parameters sometimes exhibit opposite trends, with one parameter rising while the other falls. This inconsistent variation pattern may reveal complex phenomena in neuronal dynamics.In summary, we believe that the advantage of ULIF + Time-wise compared to traditional spiking neural networks lies in the network's ability to more autonomously decide whether to fire a spike at any time point, allowing neurons to exhibit diverse dynamic characteristics.

Biological spiking neurons possess rich electrical and physical properties, and a single neuron exhibits complex spiking behavior when an external current is input. Researchers have summarized these into 20 different primary neurocomputational properties. LIF neurons possess only three neurocomputational properties: tonic spiking, Class 1 Excitability, and Integration and Coincidence Detection. However, due to its ability to control whether to fire a spike at any time point, ULIF + Time-wise can encompass other primary neurocomputational properties of biological spiking neurons that LIF neurons cannot satisfy, such as phase-locking, adaptation, and rebound spiking [46], as shown in Figure 7.

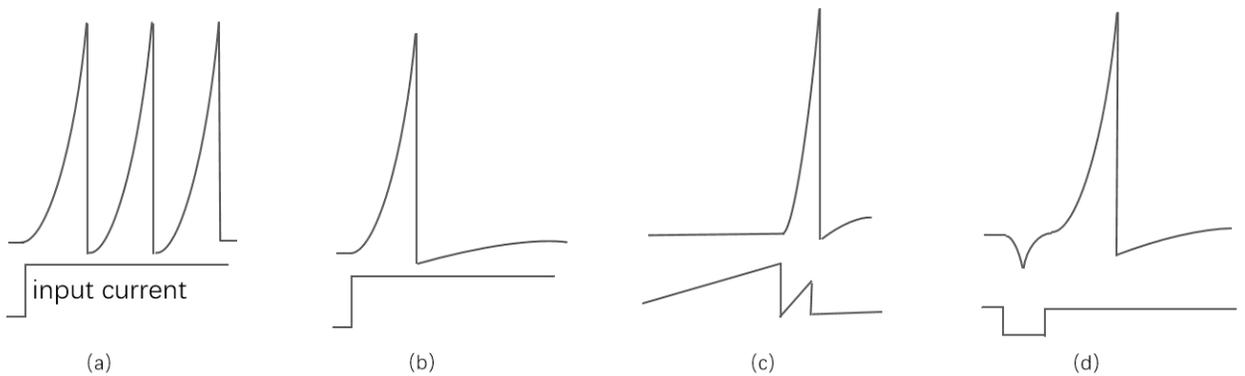

Figure 7: Four Major Neurocomputational Characteristics of Biological Spiking Neurons. Figure (a) illustrates the tonic spiking characteristic, where neurons continuously fire action potentials in the presence of sustained current. Figure (b) showcases the phasic spiking characteristic, where neurons fire a single action potential at the onset of an input current signal. Figure (c) demonstrates the accommodation characteristic, where neurons exhibit no response to high-intensity but slowly varying input currents, but generate action potentials in response to shorter duration and rapidly changing currents. Figure (d) exemplifies the rebound spike characteristic, where neurons fire action potentials at the cessation of an inhibitory input signal.

After the network training is completed, the authors observe that when using UTQS with temporal coding, the learnable parameters in the output quantization method exhibit a fixed phenomenon: the outputs of later time steps tend to have greater weights. Specifically, after learning, the network understands the characteristic that the first few time steps in temporal coding contain less information, with the amount of information increasing over time steps. Therefore, when processing data from later time steps, the network integrates information from earlier time steps, making the later time steps more decisive for the output results compared to the earlier time steps. This further demonstrates the network's effective utilization of temporal coding and maximizes the spiking neural network's ability to integrate and process spatio-temporal data.

*5.2. Analysis of Hybrid Coding*

Hybrid coding exhibits superior performance compared to direct coding and temporal coding, primarily due to its unique encoding characteristics. Ideal encoding methods should meet two core requirements: firstly, they should possess strong representational capacity for original images to minimize information loss during encoding; secondly, the generated spiking data should efficiently serve specific tasks, i.e., only retain data closely related to task demands while discarding information unrelated to task objectives.

In terms of meeting task demands, direct coding performs better in short time steps, while temporal coding excels in long time steps. From the perspective of representational capacity, temporal coding can accurately reflect the information of original images under long time steps, which direct coding cannot achieve at any time step. This is because direct coding uses a learnable convolutional layer to directly extract effective information from original images, enabling the network to rapidly obtain a large amount of spiking data required for tasks at the initial time step and effectively discard most irrelevant information in the original images. In contrast, temporal coding inputs image information into the network indiscriminately, resulting in a mixture of effective and ineffective information. In short time steps, the gap between temporal coding and direct coding is more significant due to significant information loss in temporal coding. However, in long time steps, direct coding experiences a gradual slowdown in the growth rate of effective information due to the continuous input of the same feature map, while temporal coding can continuously accumulate effective information, ultimately exhibiting superior performance compared to direct coding.

*5.3. Discussion*

Despite conducting a certain level of analysis on the dynamics of neuronal models and experimentally demonstrating the superiority of our proposed method, it remains difficult to provide a theoretical analysis for some issues. For instance, after extending the range of learnable membrane potential constants from [0, 1] to the real number domain, we found that the proportion of negative values among the learnable parameters in neurons is approximately one-sixth or less. The specific cause of this proportion remains unclear and may be attributed to dataset characteristics, network structure, or other unknown factors. Furthermore, although Time-wise has appeared in the source code of some SNN, it lacks a detailed introduction in corresponding papers. We speculate that this may be due to Time-wise not exhibiting significant effects in commonly used neuronal models. In the experimental section, Time-wise exhibited different effects in two different neuronal models, ULIF and PLIF. Although this paper has conducted a preliminary analysis of the dynamic characteristics of neurons, the fundamental cause of this phenomenon remains difficult to explain.

Additionally, the output quantization method cannot be combined with the TET method due to theoretical challenges. The TET method requires that the output result of each time step be associated with the actual label when calculating the loss function. However, in the output quantization method proposed in this paper, not all time step outputs are considered valid, and the meaning of each time step output varies. Therefore, the output result of a single time step cannot be directly associated with the actual label. The loss calculation function in TET is shown in Equation (9).

When attempting to combine the loss function in output quantization with TET, we evaluated the potential combination methods as formulated in Equation (10). but Equation (10) is not feasible in practical applications. During the network learning process, to minimize the loss, the network sets $S[t]$ to zero, resulting in the inability to perform normal gradient descent. Furthermore, the proposed method in this paper did not produce good performance when classifying neuromorphic datasets, and the specific cause of this phenomenon has not yet been identified.

## 6. Conclusion

In this paper, we used the ULIF neuronal model in conjunction with the Time-wise method and output quantization scheme to significantly improve the network's performance in processing temporal information. By alleviating the network's inherent deficiency in temporal information processing capabilities, the proposed method achieved a significant performance enhancement. In addition, we also attempted to use a hybrid encoding scheme, which enabled better performance to be obtained using temporal encoding even at lower time steps.

The method proposed in this paper has broad applicability and can be applied to any existing spiking neural networks to effectively enhance their performance. Looking ahead, this method is expected to be widely used in larger-scale neural networks and demonstrate greater potential in other computer tasks such as speech processing, image generation, and object detection.

## 7. Acknowledgment

This work supported by National Natural Science Foundation of China (No. 62176087), Natural Science Foundation of Henan (No. 242300421218), and Scientific and Technological Innovation Team of Universities in Henan Province (No. 24IRTSTHN021).

16    Author name / Neurocomputing 00 (2025) 000–000